\documentclass[11pt]{article}
\usepackage{acl-hlt2011}
\usepackage{graphicx}
\usepackage{mdwlist}
\usepackage{times}

\title{USFD at KBP 2011: Entity Linking, Slot Filling and Temporal Bounding}

\author{\textbf{Amev~Burman, Arun~Jayapal, Sathish~Kannan, Madhu~Kavilikatta} \\
	\textbf{Ayman~Alhelbawy, Leon~Derczynski, Robert~Gaizauskas}\\
	Natural Language Processing Group\\
	Department of Computer Science\\
	University of Sheffield\\
	Sheffield, S1 4DP, UK}

\date{}

\begin{document}
\maketitle


\section{Introduction}

This paper describes the University of Sheffield's entry in the 2011 TAC KBP entity linking and slot filling tasks~\cite{ji2011kbp}. We chose to participate in the monolingual entity linking task, the monolingual slot filling task and the temporal slot filling tasks. Our team consisted of five MSc students, two PhD students and one more senior academic. For the MSc students, their participation in the track formed the core of their MSc dissertation project, which they began in February 2011 and finished at the end of August 2011. None of them had any prior experience in human language technologies or machine learning before their programme started in October 2010. For the two PhD students participation was relevant to their ongoing PhD research. This team organization allowed us to muster considerable manpower without dedicated external funding and within a limited period time; but of course there were inevitable issues with co-ordination of effort and of getting up to speed. The students found participation to be an excellent and very enjoyable learning experience.

Insofar as any common theme emerges from our approaches to the three tasks it is an effort to learn from and exploit data wherever possible: in entity linking we learn thresholds for nil prediction and acquire lists of name variants from data; in slot filling we learn entity recognizers and relation extractors; in temporal slot filling we use time and event annotators that are learned from data.

The rest of this paper describes our approach and related
investigations in more detail. 
Sections~\ref{entity-linking} and~\ref{slot-filling} describe in detail
our approaches to the EL and SF tasks respectively, and Section~\ref{temporal} summarises our temporal slot filling approach. 

%
%
%
%
%
%
%
%

\section{Entity Linking Task}
\label{entity-linking}

%
%
%
%
%
%
%
%
%
%
%


\begin{figure*}
	\centering
	\includegraphics[width=.8\textwidth,height=8cm]{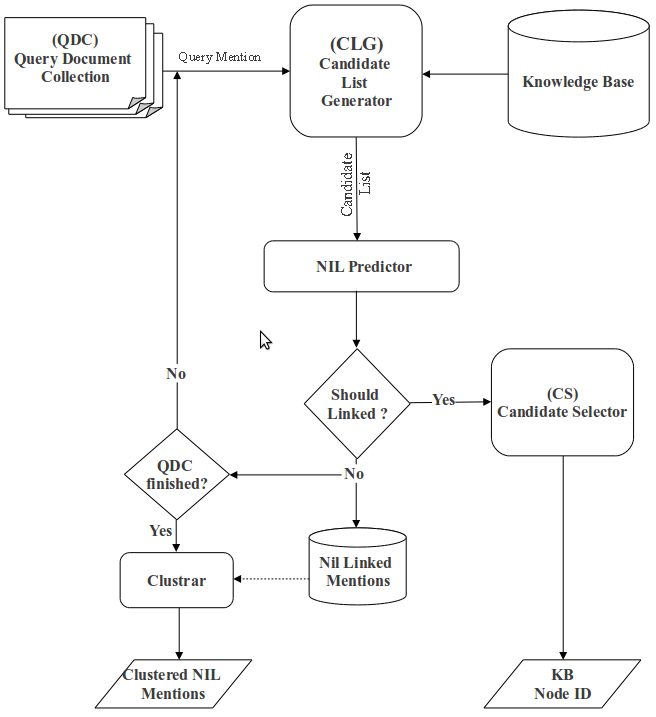}
	\caption{Flow Chart of the Initial Approach for Entity Linking}
	\label{fig:InitialApproach}
\end{figure*} 


The entity linking task is to associate a queried named entity mention, as contextualized within a given document, with a knowledge base (KB) node in a provided knowledge base which describes the same real world entity. If there is no such node the entity should be linked to Nil. There are three main challenges in this task. The first challenge is the ambiguity and multiplicity of names: the same named entity string can occur in different contexts with different meaning (e.g. \textit{Norfolk} can refer to a city in the United States or the United Kingdom); furthermore, the same named entity may be denoted using various strings, including, e.g. acronyms (\textit{USA}) and nick names (\textit{Uncle Sam}). The second challenge is that the queried named entity may not be  found in the knowledge base at all. The final challenge is to cluster all Nil linked mentions. 

\subsection{System Processing}
Our system consists of four stage model, as shown in Figure \ref{fig:InitialApproach}:
\begin{enumerate*}
\item Candidate Generation: In this stage, all KB nodes which might possibly be linked to the query entity  are retrieved.
\item Nil Predictor: In this stage, a binary classifier is applied to decide whether the query mention should be linked to a KB node or not.
\item Candidate Selection: In this stage, for each query mention that is to  be linked to the KB, one candidate from the candidate set is selected as the link for the query mention.
\item Nil Mention Clustering: In this stage, all Nil linked query mentions are clustered so that each cluster contains all mentions that should be linked to a single KB node, i.e. pertain to the same entity.
\end{enumerate*}

\subsubsection{Candidate Generation}\label{sec:candidate_generation}
The main objective of the candidate generation process is to reduce the search space of potential link targets from the full KB to a small subset of plausible candidate nodes within it. The query mention is used, both as a single phrase and as the set of its constituent tokens, to search for the query string in the titles and body text of the KB node. 

\paragraph{Variant name extraction}
We extracted different name forms for the same named entity mention from a Wikipedia dump. Hyper-links, redirect pages and disambiguation pages are used to associate different named entity mentions with the same entity~\cite{LNESKB2010,varma2009iiit}. This repository of suggested name variants is then used in query expansion to extend the queries regarding a given entity to all of its possible names. Since the mention of the entity is not yet disambiguated, it is not necessary for all suggested name variants to be accurate.

\paragraph{Query Generation}
We generated sets of queries according to two different strategies. The first strategy is based on name variants, using the previously built repository of Wikipedia name variants. The second strategy uses additional named entity (NE) mentions for query expansion: the Stanford NE recognizer~\cite{finkel2005incorporating} is used to find NE mentions in the query document, and generates a query containing the query entity mention plus all the NE mentions found in the query document, 

\begin{table*}
\centering 		
\begin{tabular}{|c|c|c|c|c|c|}
\hline \textbf{Data Set} & \textbf{$\alpha$} & \textbf{Precision} & \textbf{Recall} & \textbf{F1} & \textbf{Accuracy}\\
\hline TAC 2009 & 5.9 & 59.42 & 80.18 & 68.26 & 68.01\\
\hline TAC 2010 & 5.9 & 75.36 & 77.65 & 76.48 & 78.36\\
\hline TAC 2011 & 5.9 & 74.83 & 66.90 & 70.64 & 72.22\\
\hline
\end{tabular}
\caption{Performance of Nil Predictor using highest score candidate}
\label{tab:Initial_Classification1}
\end{table*}

\begin{table*}
\centering 		
\begin{tabular}{|c|c|c|c|c|c|}
\hline \textbf{Data Set} & \textbf{$\beta$} & \textbf{Precision} & \textbf{Recall} & \textbf{F1} & \textbf{Accuracy}\\
\hline TAC 2009 & 0.16 & 54.74 & 64.84 & 59.36 & 61.91\\
\hline TAC 2010 & 0.16 & 57.92 & 62.35 & 60.06 & 62.40\\
\hline TAC 2011 & 0.16 & 54.69 & 31.14 & 39.68 & 52.71\\
\hline
\end{tabular}
\caption{Performance of Nil Predictor using difference between two highest scored candidates}
\label{tab:Initial_Classification2}
\end{table*}

\paragraph{Retrieval}
After query generation, we performed document retrieval using Lucene. All knowledge base nodes, titles, and wiki-text were included in the Lucene index. Documents are represented as in the Vector Space Model (VSM). For ranking results, we use the default Lucene similarity function which is closely related to cosine similarity .

\subsubsection{Nil Prediction}\label{sec:intial_NIL_Predection}
In many cases, a named entity mention is not expected to appear in the knowledge base. We need to detect these cases and mark them with a \texttt{NIL} link. The \texttt{NIL} link is assigned after generating a candidate list (see Varma et al.~\shortcite{varma2009iiit}, Radford et al.~\shortcite{radforddocument}). 

If the generated candidate list is empty, then the query mention is linked to \texttt{NIL}.  If the candidate list is not empty, we use two techniques to find a candidate. The first just chooses the highest ranked candidate, i.e. the highest scoring candidate using the Lucene similarity score. If the the highest scoring candidate score is above some threshold $ \alpha $ then the candidate is selected and if it is under the threshold, the predictor links the mention to \texttt{NIL}. The second technique calculates the difference between the  scores of two highest scoring candidates, then compares this difference with some threshold $ \beta $; if the difference exceeds the threshold the highest scoring candidate is selected, otherwise the query mention is linked to  \texttt{NIL}.. Results of these two techniques are shown in Tables \ref{tab:Initial_Classification1} and \ref{tab:Initial_Classification2}.

\begin{table*}
\begin{center} 		
\begin{tabular}{|c|c|c|c|c|}
\hline \textbf{Levenshtein distance}& \textbf{RI} & \textbf{Precision} & \textbf{Recall} & \textbf{F1}\\
\hline 0 & 99.98 & 95.06 & 99.15 & 97.06\\
\hline 1 & 99.98 & 92.22 & 99.21 & 95.59\\
\hline
\end{tabular}
\caption{Performance of Nil Clustering}
\label{Tab:Clustering_Result}
\end{center}
\end{table*}

\paragraph{Parameter Setting for Nil Matching}
To find the best thresholds, a Na\"ive Bayes classifier is trained using the TAC 2010 training data. We created training data as follows. For each query in the training set, we generate a candidate list and the highest scoring document is used as a feature vector. If it is the correct candidate then the output is set to true else the output set to false. From this set of instances a classifier is learned to get the best threshold.

\subsubsection{Candidate Selection}\label{sec:candidate_selection}
The candidate selection stage will run only on a non-empty candidate list, since an empty candidate list means linking the query mention to \texttt{NIL}. For each query, the highest-scoring candidate is selected as the correct candidate. 

\subsubsection{Nil Clustering}
A simple clustering technique is applied. The Levenshtein distance is measured between the different mentions and if the distance is under a threshold $\alpha$, the mentions are grouped into the same cluster. Two experiments are carried out and results are presented in Table \ref{Tab:Clustering_Result}. As shown clustering according to the string equality  achieves better results than allowing a distance of one.

\paragraph{Data Set:}
The TAC2011 data set contains 2250 instances of which 1126 must be linked to ``Nil". In the gold standard, the 1126 Nil instances are clustered into 814 clusters. Only those 1126 instances are sent to the clustering module to check its performance separately, regardless of Nil predictor performance.

\paragraph{Evaluation Metric:}
``All Pair Counting Measures" are used to evaluate the similarity between two clustering algorithm's results. This metric examines how likely the algorithms  are to group or separate a pair of data points together in different clusters. These measures are able to compare clusterings with different numbers of clusters.

The Rand index \cite{rand1971objective} computes similarity between the system output clusters (output of the clustering algorithm) and the clusters found in a gold standard. So, the Rand index measures the percentage of correct decisions -- pairs of data points that are clustered together in both system output and gold standard, or, clustered in different clusters in both system output and gold standard -- made by the algorithm. It can be computed using the following formula:

$$RI=\frac{Tp+Tn}{Tp+Tn+Fp+Fn}$$

\subsection{Evaluation}
In this section we provide a short description of different runs and their results. All experiments are evaluated using the B-Cubed$^{+}$ and micro average scoring metrics. In our experimental setup, a threshold $\alpha=5.9$ is used in Nil-Predictor and Levenshtein distance = 0 is used for Nil clustering. The standard scorer released by the TAC organizers is used to evaluate each run, with results in Table \ref{tab:TAC2011RunsResults}. Different query schemes are used in different runs as follows. 
\begin{enumerate*}
\item Wiki-text is not used, with search limited to nodes titles only. The search scheme used in this run uses query mention only.
\item Wiki-text is used. The search scheme used in this run uses the query mention and the different name variants for the query mention.
\item Wiki-text is used, The search scheme used in this run uses the query mention and the different name variants for the query mention. Also, it uses the query document named entities recognized by the NER system to search within the wiki-text of the node.
\end{enumerate*}

\begin{table*}
\centering 		
\begin{tabular}{|c|c|c|c|c|}
\hline \textbf{Run}  & \textbf{Micro Average} &\textbf{B{\textasciicircum}3 Precision} & \textbf{B{\textasciicircum}3 Recall} & \textbf{B{\textasciicircum}3 F1}\\
\hline 1 & 49.2 & 46.2 & 48.8 & 47.5 \\
\hline 2 & 43.6 & 40.8 & 43.2 & 42.0 \\
\hline 3 & 46.8 & 44.0 & 45.6 & 44.8 \\
\hline 
\end{tabular}
\caption{Results of Three runs for entity linking task }
\label{tab:TAC2011RunsResults}
\end{table*}

\section{Slot Filling Task}
\label{slot-filling}

There are a number of different features that can describe an entity. For an organisation, one might talk about its leaders, its size, and place of origin. For a person, one might talk about their gender, their age, or their religious alignment. These feature types can be seen as `slots', the values of which can be used to describe an entity.

The slot-filling task is to find values for a set of slots for each of a given list of entities, based on a knowledge base of structured data and a source collection of millions of documents of unstructured text. In this section, we discuss our approach to slot filling.

%
%
%

\subsection{System Processing}

Our system is structured as a pipeline. For each entity/slot pair, we begin by selecting documents that are likely to bear slot values, using query formulation (Section~\ref{qf-step}) and then information retrieval (Section~\ref{ir-step}) steps. After this, we examine the top ranking returned texts and, using  learned classifiers, attempt to extract all standard named entity mentions plus mentions of other entity types that can occur as slot values (Section~\ref{entity-step}). Then we run a learned slot-specific relation extractor over the sentences containing an occurrence of the target entity and an entity of the type required as a value for the queried slot, yielding a
list of candidate slot values (Section~\ref{csve-step}). We then rank these candidate slot values and return a slot value, or list of slot values in the case of list-valued slots, from the best candidates (Section~\ref{slot-value-selection}).

\subsubsection{Preprocessing and Indexing}
\label{ir-step}

Information Retrieval (IR) was used to address the tasks of Slot Filling (SF) and Entity Linking (EL) primarily because it helps in choosing the right set of documents and hence reduces the number of documents that need to be processed further down the pipeline. Two variations of IR were used in the SF task: document retrieval (DR) and passage retrieval (PR).

The documents were parsed to extract text and their parent elements using JDOM and then indexed using Lucene. We used Lucene's standard analyzer for indexing and stopword removal. The parent element of the text is used as field name. This gives the flexibility of searching the document using fields and document structure as well as just body~\cite{baeza1999modern}. Instead of returning the text of the document, the pointers or paths of the document were returned when a search is performed. For searching and ranking, Lucene's default settings were used.

For passage retrieval, various design choices were considered~\cite{Rob04a} and a two stage process was selected. In the two stage process, the original index built for DR is used to retrieve the  top $n$ documents and the plain text (any text between two SGML elements) is extracted as a separate passage. A temporary mini-index is then built on the fly  from these passages. From the temporary index, the top $n$ \emph{passages} are retrieved for a given query. Instead of returning the text of the passages, the location of the passage (element retrieval) in the document is returned as a passage offset within a  document referenced by a file system pointer. Two versions of passage systems were created, one that removes stop-words while indexing and searching and other that keeps the stop words. For ranking, Lucene's default settings were used.

Finally the IR system and the query formulation strategies were evaluated on the DR task to determine the optimal number of top ranked documents to retrieve for further processing down the pipeline and for PR. This evaluation is further discussed in Section~\ref{subsec:slotfill-eval} below.

\subsubsection{Query Formulation}
\label{qf-step}

This step generates a query for the IR system that attempts to retrieve the best documents for a given entity and slot type.

\paragraph{Variant name extraction}
\label{variant}

Variant names are the alternate names of an entity (persons or organizations only for the slot filling task in 2011) which are different from their formal name. These include various name forms such as stage names, nick names and abbreviations. Many people have an alias; some people even have more than one alias. In several cases people are better known to the world by their alias names rather than their original name. For example, Tom Cruise is well known to the world as an actor, but his original name is Thomas Cruise Mapother IV. Alias names are very helpful to disambiguate the named entity, but in some cases the alias names are also shared among multiple people. For example, MJ is the alias name for both Michael Jackson (Pop Singer) and Michael Jordan (Basketball player).	

Variant name forms are used for query formulation. 
The methods used in the slot filling task for extracting variant name forms from a Wikipedia page are:

\begin{itemize*}
\item Extract all the name attributes from the infobox, such as nickname, birth name, stage name and alias name.
\item Extract the title and all bold text from the first paragraph of the article page.
\item Extract the abbreviations of the entity name by finding patterns like ``\texttt{(ABC)}" consisting of all capital letters appearing after the given entity name. For example, TCS is an abbreviation of the entity Tata Consultancy Service in case of the following pattern \emph{Tata Consultancy Service, (TCS)}.
\item Extract all redirect names that refer to the given entity. For example, the name `King of Pop' automatically redirects to the entity named `Michael Jackson'. 
\item In the case of ambiguous names extract all the possible entity names that share the same given name from the disambiguation page. 
\end{itemize*}

A variant name dictionary was created by applying all the above methods to every entity in the Wikipedia dump. Each line of the dictionary contains the entity article title name as in Wikipedia followed by one of the variant name forms. This dictionary is then used  at query time to find the variant name forms of the given entity.

\paragraph{Slot keyword collection}

The query formulation stage deals with developing a query to retrieve the relevant documents or passages for each slot of each entity. 
Our approach is as follows:

\begin{enumerate*}
\item Collect manually (by referring to public sources such as Wikipedia) a list of keywords for each slot query. Some example keywords for the per:countries\_of\_residence slot query are `house in', `occupies', `lodges in', `resides in', `home in', `grew up in' and `brought up in'.
\item Extract all the alternate names of the given entity name the variant name dictionary (Section~\ref{variant}).
\item Formulate a query for each slot of an entity by including terms for entity mention, variant names and keywords collected for the slot query in the first step. These terms are interconnected by using Boolean operators.
\item The formulated query is then fed into the IR component and the top $n$ documents retrieved.
\end{enumerate*}

\subsubsection{Entity Identification}
\label{entity-step}

Given the top $n$ documents returned by the previous phase of the system, the next task is to identify potential slot values. To do this we used entity recognizers trained over existing annotated datasets plus some additional datasets we developed. For a few awkward slot value types we developed regular expression based matchers to identify candidate slot fills. We have also developed a restricted coreference algorithm for identifying coreferring entity mentions, particularly mentions coreferring with the query (target) entity,

%

\paragraph{Named Entity Recognition}


The Stanford Named Entity Recognition (NER) tool ~\cite{finkel2005incorporating} was used to find named entities. It is  a supervised learning conditional random field based approach which  comes with a pre-trained model for three entity classes. Because we needed a broader range of entity classes we re-trained the classifier using the MUC6 and MUC7 datasets~\footnote{LDC refs. LDC2001T02, LDC2003T13} and NLTK~\cite{Bir09} gazetteers. Training the classifier was not straightforward as the source data had to be reformatted into the format recognized by Stanford NER. The MUC datasets provided training data for the entities Location, Person, Organization, Time, Person, Money, Percent, Date, Number and Ordinal. More classes were added to the MUC training dataset since the slot-filling task required nationality, religion, country, state, city and cause-of-death slot fill types to be tagged as well. For country, state and city, which can be viewed as sub-types of type location we semi-automatically adapted the MUC training data by finding all location entities in the data, looking them up in a gazetteer and then manually adding their sub-type. For nationalities, causes of death and religion, we extracted lists of nationalities, causes of death and religions from Wikipedia. In the case of nationality and causes of death we searched for instances of these in the MUC data and then labelled them to provide training data. For religion, however, because there were so few instances in the MUC corpus and because of issues in training directly on Wikipedia text, we used a post-classifier list matching technique to identify religions. 

The trained classifier was  used to identify and tag all  mentions of the  entity types it knew about in the documents and/or passages returned by the search engine. These tagged documents were then passed on to the co-reference resolution system. After some analysis we discovered that in some cases the target entity supplied in the quey was not being correctly tagged by the entity tagger. Therefore we added a final phase to our entity identifier in which all occurrences of the target entity were identified and tagged with the correct type, regardless of whether they had or had not been tagged correctly by the CRF entity tagger. s

\paragraph{Restricted Co-reference Resolution}

To identify the correct slot fill for an entity requires not just identifying mentions which are of the correct slot fill type but of ensuring that the mention stands in the appropriate relation to the target entity -- so, to find Whistler's mother requires not only finding entities of type PERSON, but also determining that the person found stands the relation ``mother-of" to Whistler. Our approach to relation identification, described in the next section, relies on the relation being expressed in a sentence in which both the candidate slot fill and the target entity occur. However, since  references to the target entity or to the slot fill may be anaphoric, ability to perform coreference resolution is required.

Off-the-shelf co-reference resolvers, such as the Stanford CRF-based coreference tool,  proved too slow to complete slot-filling runs in a reasonable timeframe. Therefore, we designed a custom algorithm to do limited heuristic coreference  to suit the slot-filling task. Our algorithm is limited in two ways. First, it only considers coreferring references to the target entity and ignores any coreference to candidate slot fills or between any other entities in the text. Second, only a  limited set of anaphors is considered.  In the case of target entities of type PERSON the only anaphors considered are personal and possessive pronouns such as \emph{he}, \emph{she}, \emph{his} and \emph{her}.  In these cases it also helps to identify whether the target entity is male or female. We trained the maximum entropy classifier provided with NLTK with a list of male names and female names also from NLTK. The last and second to last characters for each name were taken as features for training the classifier. Based on the output produced by the classifier, the system decides whether certain pronouns are candidate anaphors for resolving with the target entity. For example, when the output produced by the classifier for the PERSON entity {\it Michael Jackson} is male, only mentions of  \textit{he} and \textit{his} will be considered as candidate anaphors.

When the target entity is of type ORGANIZATION, only the pronoun \textit{it} or common nouns referring to types of organization,  such as \textit{company}, \textit{club}, \textit{society}, \textit{guild}, \textit{association}, etc. are considered as potential anaphors. A list of such organization nouns is extracted from GATE.


For both PERSONs and ORGANIZATIONs, when candidate anaphors are identified the algorithm resolves them to the target entity if a tagged mention of the target entity is the textually closest preceding tagged mention of an entity of the target entity type. For example, {\it he} will be coreferred with {\it Michael Jackson} if a tagged instance of {\it Michael Jackson},  or something determined to corefer to it, is the closest preceding mention of a male entity of type PERSON. If an intervening male person is found, then no coreference link is made. When coreference is established, the anaphor -- either pronoun or common noun -- is labelled as `target entity".

This approach to coreference massively reduces the complexity of the generalized coreference task, making it computationally tractable within the inner loop of processing multiple documents per slot per target entity. Informal evaluation across a small number of manually examined documents showed the algorithm performed quite well. 

\subsubsection{Candidate Slot Value Extraction}
\label{csve-step}

The next sub-task is to extract candidate slot fills by determining if the appropriate relation holds between a mention of the target entity and a mention of an entity of the appropriate type for the slot. For example if the slot is \texttt{date\_of\_birth}  and the target entity is {\it Michael Jackson} then does the \texttt{date\_of\_birth} relation hold between some textual mention of the target entity {\it Michael Jackson} (potentially an anaphor labelled as target entity) and some textual mention of an entity tagged as type DATE.

The general approach we took  was to select all sentences that contained both a target entity mention as well as a mention of the slot value type  and run a binary relation detection classifier to detect relations between every potentially related target entity mention-slot value type mention in the sentence. If the given relation is detected in the sentence, the slot value for the relation (e.g. the entity string) is identified as a candidate value for the slot of the target entity.

\begin{table*}
\centering
\begin{tabular}{ | c | c | c | c | c | c | l | }
\hline
\textbf{Run} & \textbf{Recall} & \textbf{Precision} & \textbf{F1} & \textbf{Retrieval} & \textbf{Co-ref?} & \textbf{Slot extractor} \\
\hline
1 & 1.38\% & 2.43\% & 0.0176 & document & no & BoW \\
2 & 5.08\% & 4.84\% & 0.0496 & document & yes & BoW + ngram \\
3 & 1.16\% & 2.97\% & 0.0167 & passage & yes & BoW + ngram \\
\hline
\end{tabular}
\caption{Slot filling results for USFD2011.}
\label{tab:sf-results}
\end{table*}

\paragraph{Training the Classifiers}

A binary relation detection classifier needed to be trained for each type of slot. Since there is no  data explicitly labelled with these relations we used a distant supervision approach (see, e.g., \newcite{Min09}). This relied on an external knowledge base -- the infoboxes from Wikipedia -- to help train the classifiers. In this approach, the fact names from the Wikipedia infoboxes were mapped to the KBP. These known slot value pairs from the external knowledge base were used to extract sentences that contain the target entity and the known slot value. These formed positive instances. Negative instances were formed from sentences containing the target entity and an entity mention of the appropriate type for the slot fill, but whose value did not match the value taken from the infobox (e.g. a DATE, but not the date of birth as specified in the infobox for the target entity). The classifiers learned from this data were then used on unknown data to extract slot value pairs. 

\paragraph{Feature Set}
Once the positive and negative training sentences were extracted, the next step was to extract feature sets from these sentences which would then be used by machine learning algorithms to train the classifiers. Simple lexical features and surface features were included in the feature set. Some of the features used include:

\begin{itemize*}
\item Bag of Words: all words in the training data not tagged as entities were used as binary features whose value is 1 or 0 for the instance depending on whether they occur in sentence from which the training instance is drawn.
\item Words in Window: like Bag of Words but only words between the target entity and candidate slot value mentions plus two words before and after are taken as features.
\item N-grams: like bag of words, but using bi-grams instead of unigrams
\item Token distance: one of three values -- short ($<=3$), medium ($>3$ and $<=6$) or long ($>6$) -- depending on the distance in tokens between the  the target entity and candidate slot value mentions.
\item Entity in between: binary feature indicating whether there is another entity of the same type between the candidate slot value mention and the target entity.
\item Target first: binary feature indicating whether the target entity comes before the candidate slot value in the sentence?
\end{itemize*}

We experimented with both the Naive Bayes and Maximum Entropy classifiers in the NLTK. For technical reasons could not get the maximum entropy classifier working in time for the official test runs, so our submitted runs used the Naive Bayes classifiers, which is almost certainly non-optimal given the non-independence of the features.

\subsubsection{Slot Value Selection}
\label{slot-value-selection}

The final stage in our system is to select which candidate slot value (or slot values in the case of list-valued slots) to return as the correct answer from the candidate slot values extracted by the relation extractor in the previous stage. To do this
we rank the candidates identified in the candidate slot value extraction stage. Two factors are considered in ranking the candidates: (1) the number of times a value has been extracted, and (2) the confidence score provided for each candidate by the relation extractor classifier. 
If any value in the list of possible slot values occurs more than three times, then the system uses the number of occurrences as a ranking factor. Otherwise, the system uses the confidence score as a ranking factor. In the first case candidate slot values are sorted on the basis of number of occurrences. In the second case values are sorted on the basis of confidence score. In both cases the top n value(s) are taken as the correct slot value(s) for the given slot query. We use $n=1$ for single-valued slots $n=3$ for list-valued slots. 

Once the system selects the final slot value(s), the final results are written to a file in the format required by the TAC guidelines.

\subsection{Evaluation}
\label{subsec:slotfill-eval}

We evaluated both overall slot-filling performance, and also the performance of our query formulation / IR components in providing suitable data for slot-filling.

\subsubsection{Overall}

We submitted three runs: one with document-level retrieval, no coreference resolution, and bag-of-words extractor features; a second with document-level retrieval, coreference resolution, and n-gram features; a third with passage-level retrieval, coreference resolution, and n-gram features. Our results are in Table~\ref{tab:sf-results}.

\subsubsection{Query Formulation/Document Retrieval Evaluation}


\begin{table}
\small
\begin{tabular}{|c|r|r|r|r|r|r|}
\hline
\textbf{Slots}	& \textbf{TD} 	& \textbf{NQ} 	& \textbf{LC} 		& \textbf{SC} 		& \textbf{LR} 		& \textbf{SR} \\
\hline
All 	& 5 	& 742 	& 0.468 	& 0.252 	& 0.954 	& 0.252 \\ 
All 	& 10 	& 742 	& 0.534 	& 0.307 	& 1.558 	& 0.307 \\
All 	& 20 	& 742 	& 0.589 	& 0.358 	& 2.434 	& 0.358 \\
All 	& 50 	& 742 	& 0.616 	& 0.391 	& 3.915 	& 0.391 \\
\hline
\end{tabular}
\caption{Coverage and Redundancy Analysis for All Entities and All Slots. 
*TD = Top Docs, NQ = No of Queries, LC = Lenient Coverage, SC= Strict Coverage, LR = Lenient Redundancy and SR = Strict Redundancy.}
\label{tab:cov-redund}
\end{table}

We evaluated query formulation and document retrieval using the coverage and redundancy measures introduced by \newcite{Rob04a}, originally developed for question answering. Coverage is the proportion of questions for which answers can be found from the documents or passages retrieved, while redundancy is the average number of documents or passages retrieved which contain answers for a given question or query. These measures may be directly carried over to the slot filling task, where we treat each slot as a question. 
 
The evaluation used the 2010 TAC-KBP data for all entities and slots; results are shown in Table~\ref{tab:cov-redund}. Strict and lenient versions of each measure were used, where for the strict measure both document ID and response string must match those in the gold standard, while for the lenient only the response string must match, i.e. the slot fill must be correct but the document in which it is found need not be one which has been judged to contain a correct slot fill. This follows the original strict and lenient measures implemented in the tool we used to assist evaluation, IR4QA~\cite{sanka2005}.

The results table shows a clear increase in all measures as the number of top ranked documents is increased.
With the exception of lenient redundancy, the improvement in the scores from the top 20 to the 50 documents is not very big. Furthermore if 50 documents are processing through the entire system as opposed to 20, the additional 30 documents  will both more than double processing times per slot and introduce many more potential distractors for the correct slot fill (See Section~\ref{slot-value-selection}). For these reasons we chose to limit the number of documents passed on from this stage in the processing to 20 per slot. Note that this bounds our slot fill performance to just under 60\%.

\subsubsection{Entity Extraction and Coreference Evaluation}

We evaluated our entity extractor as follows.  We selected one entity and one slot for entities of type ORGANIZATION and one for entities of type PERSON and gathered the top 20 documents returned by our query formulation and document retrieval system for each of these entity-slot pairs. We manually annotated all candidate slot value across these two twenty document sets to provide a small  gold standard test set. For candidate slot  fills in documents matching the ORGANIZATION query, overall F-measure for the entity identifier was 78.3\% while for candidate slot  fills in documents matching the PERSON query, overall F-measure for the entity identifier was 91.07\%. We also manually evaluated our coreference approach over the same two document sets and arrived at an F-measure of  73.07\% for coreference relating to the ORGANIZATION target entity and 90.71\% for  coreference relating to the PERSON target entity. We are still analyzing the wide difference in performance of both entity tagger and coreference resolver when processing documents returned in response to an ORGANIZATION query as compared to documents returned in response to a  PERSON query. 

%

\subsubsection{Candidate Slot Value Extraction Evaluation}

\begin{table}
\small
\begin{center}
\begin{tabular}{|l|l|l|l|l|}
\hline
  & DATE & PER & LOC & ORG\\
\hline
 +ve  & 97.5 & 95 & 92.5 & 83.33\\
-ve   & 87.5 & 95 & 97.5& 85\\
\hline
\end{tabular}
\caption{Estimated \% Training Instances Correct}
\label{tab:training-data-eval}
\end{center}
\end{table}

\begin{table}
\small
\begin{tabular}{|l|l|l|l|l|}
\hline
  Dataset& DATE & PER & LOC & ORG\\
\hline
Training &  82.34 & 78.44 & 66.29 & 80\\
Handpicked +ve    & 62.5 & 40 & 36.37 & 45.45\\
Handpicked -ve   & 100 & 75 & 71.42& 88.89\\
\hline
\end{tabular}
\caption{\% Slot Values  Correctly Extracted}
\label{tab:model-eval}
\end{table}

To evaluate our candidate slot value extraction process we did two separate things. First we assessed the quality of training data provided by our distant supervision approach. Since it was impossible to check all the training data produced manually we randomly sampled 40 positive examples for each of four slot types (slots expecting DATEs, PERSONs, LOCATIONs and ORGANIZATIONs) and 40 negative examples for each of four slot types. Results of this evaluation are in Table \ref{tab:training-data-eval}.

In addition to evaluating the quality of the training data we generated, we did some evaluation to determine the optimal feature set combination. Ten fold cross validation figures for the optimal feature set over the training data are shown in the first row in Table \ref{tab:model-eval}, again for a selection of one slots from each of four slot types . Finally we evaluated the slot value extraction capabilities on a small test set of example sentences selected from the source collection to ensure they contained the target entity and the correct answer, as well as some negative instances,  and manually processed to correctly annotate the entities within them (simulating perfect upstream performance). Results are shown in rows2 and 3 of Table \ref{tab:model-eval}. The large difference in performance between the ten fold cross validation figures over the training and the evaluation against the  small handpicked and annotated gold standard from the source collection may be due to the fact that the training data was Wikipedia texts while the test set is news texts and potentially other text types such as blogs; however, the handpicked test set is very small (70 sentences total) so generalizations may not be warranted. 

\section{Temporal Filling Task}
\label{temporal}

The task is to detect upper and lower bounds on the start and end times of a state denoted by an entity-relation-filler triple. This results in four dates for each unique filler value. There are two temporal tasks, a full temporal bounding task and a diagnostic temporal bounding task. We provide the filler values for the full task, and TAC provides the filler values and source document for the diagnostic task. Our temporal component did not differentiate between the two tasks; for the full task, we used output values generated by our slot-filling component. 

We approached this task by annotating source documents in TimeML~\cite{pustejovsky2003timeml}, a modern standard for temporal semantic annotation. This involved a mixture of off-the-shelf components and custom code. After annotating the document, we attempted to identify the TimeML event that best corresponded to the entity-relation-filler triple, and then proceeded to detect absolute temporal bounds for this event using TimeML temporal relations and temporal expressions. We reasoned about the responses gathered by this exercise to generate a date quadruple as required by the task.

In this section, we describe our approach to temporal filling and evaluate its performance, with subsequent failure analysis.

\subsection{System Processing}

We divide our processing into three parts: initial annotation, selection of an event corresponding to the persistence of the filler's value, and temporal reasoning to detect start and finish bounds of that state.

\subsubsection{TimeML Annotation}

Our system must output absolute times, and so we are interested in temporal expressions in text, or TIMEX3 as they are in TimeML. We are also interested in events, as these may signify the start, end or whole persistence of a triple. Finally we need to be able to determine the nature of the relation between these times and events; TimeML uses TLINKs to annotate these relationships.

We used a recent version of HeidelTime~\cite{strotgen2010heideltime} to create TimeML-compliant temporal expression (or \textbf{timex}) annotations on the selected document. This required a document creation time (\textbf{DCT}) reference to function best. For this, we built a regular-expression based DCT extractor\footnote{https://bitbucket.org/leondz/add-dct} and used it to create a DCT database of every document in the source collection (this failed for one of the 1~777~888 documents; upon manual examination, the culprit contained no hints of its creation time).

The only off-the-shelf TimeML event annotation tool found was Evita~\cite{sauri2005evita}, which requires some preprocessing. Specifically, explicit sentence tokenisation, verb group and noun group annotations need to be added. For our system we used the version of Evita bundled with TARSQI\footnote{http://timeml.org/site/tarsqi/toolkit/}. Documents were preprocessed with the ANNIE VP Chunker in GATE\footnote{http://gate.ac.uk/}. We annotated the resulting documents with Evita, and then stripped the data out, leaving only TimeML events and the timexes from the previous step.

At this point, we loaded our documents into a temporal annotation analysis tool, CAVaT~\cite{derczynski2010analysing}, to simplify access to annotations. Our remaining task is temporal relation annotation. We divided the classes of entity that may be linked into two sets, as per TempEval~\cite{verhagen2010semeval}: intra-sentence event-time links, and inter-sentence event-event links with a 3-sentence window. Then, two classifiers were learned for these types of relation using the TimeBank corpus\footnote{LDC catalogue entry LDC2006T08.} as training data and the linguistic tools and classifiers in NLTK\footnote{http://www.nltk.org/}. Our feature set was the same used as Mani et al.~\shortcite{mani2007three} which relied on surface data available from any TimeML annotation.

\subsubsection{Event Selection}

To find the timexes that temporally bound a triple, we should first find events that occur during that triple's persistence. We call this task ``event selection". Our approach was simple. In the first instance we looked for a TimeML event whose text matched the filler. Failing that, we looked for sentences containing the filler, and chose an event in the same sentence. If none were found, we took the entire document text and tried to match a simplified version of the filler text anywhere in the document; we then returned the closest event to any mention. Finally, we tried to find the closest timex to the filler text. If there was still nothing, we gave up on the slot.

\begin{table}
\begin{center}
\begin{tabular}{| l | r |}
\hline
\textbf{Slot name} & \textbf{Count} \\
\hline
per:title &73\\
per:member\_of &36\\
per:employee\_of &33\\
org:subsidiaries &29\\
per:schools\_attended& 17\\
per:cities\_of\_residence &14\\
per:stateorprovinces\_of\_residence& 13\\
per:spouse &11\\
per:countries\_of\_residence& 11\\
org:top\_members/employees &10\\
\hline
Total & 247 \\
\hline
\end{tabular}
\caption{Distribution of slot types in the available training and sample data.}
\label{tab:training-slots}
\end{center}
\end{table}

\subsubsection{Temporal Reasoning}

Given a TimeML annotation and an event, our task is now to find which timexs exist immediately before and after the event. We can detect these by exploiting the commutative and transitive nature of some types of temporal relation. To ensure that as many relations as possible are created between events and times, we perform pointwise temporal closure over the initial automatic annotation with CAVaT's consistency tool, ignoring inconsistent configurations. Generating temporal closures is computationally intensive. We reduced the size of the dataset to be processed by generating isolated groups of related events and times with CAVaT's \texttt{subgraph} modules, and then computed the closure over just these ``nodes".

We now have an event representing the slot filler value, and a directed graph of temporal relations connecting it to times and events, which have been decomposed into start and end points. We populate the times as follows:

\begin{itemize*}
\item $T_1$: Latest timex before event start
\item $T_2$: Earliest timex after event start
\item $T_3$: Latest timex before event termination
\item $T_4$: Earliest timex after event termination
\end{itemize*}

\begin{table}
\begin{center}
\begin{tabular}{| l | c |}
\hline
\textbf{Slot Type} & \textbf{Score} \\
\hline
per:stateorprovinces\_of\_residence&	0.583 \\
per:employee\_of&	0.456\\
per:countries\_of\_residence&	0.787\\
per:member\_of	&0.534\\
per:title&	0.529\\
org:top\_members/employees	&0.571	\\
per:spouse&	0.535\\
per:cities\_of\_residence&	0.744\\
\hline
Weighted overall score & 0.552 \\
\hline
\end{tabular}
\caption{Scores of our temporal system over the union of temporal sample and training data.}
\label{tab:temporal_scorer_training}
\end{center}
\end{table}

Timex bounds are simply the start and end points of an annotated TIMEX3 interval. We resolve these to calendar references that specify dates in cases where their granularity is greater than one day; for example, using 2006-06-01 and 2006-06-30 for the start and end of a \texttt{2006-06} timex. Arbitrary points are used for season bounds, which assume four seasons of three months each, all in the northern hemisphere. If no bound is found in the direction that we are looking, we leave that value blank.

\subsection{Evaluation}

Testing and sample data were available for the temporal tasks\footnote{LDC catalogue entries LDC2011E47 and LDC2011E49.}. These include query sets, temporal slot annotations, and a linking file describing which timexes were deemed related to fillers. The distribution of slots in this data is given in Table~\ref{tab:training-slots}. To test system efficacy we evaluated output performance with the provided entity query sets against these temporal slot annotations. Results are in Table~\ref{tab:temporal_scorer_training}, including per-slot performance.

\begin{table}
\begin{center}
\begin{tabular}{|l|c|c|c|}
\hline
\textbf{Retrieval level} & \textbf{Precision} & \textbf{Recall} & \textbf{F1} \\
\hline
Document & 1.52\% & 0.70\% & 0.96\% \\
Paragraph & 0.14\% & 0.79\% & 0.24\% \\
\hline
\end{tabular}
\caption{Full temporal slot-filling results}
\label{tab:temporal-slot-filling}
\end{center}
\end{table}

Results for the full slot-filling task are given in Table~\ref{tab:temporal-slot-filling}. This relies on accurate slot values as well as temporal bounding. An analysis of our approach to the diagnostic temporal task, perhaps using a corpus such as TimeBank, remains for future work.

\section{Conclusion}
\label{conclusion}

We set out to build a framework for experimentation with knowledge base population. This framework was created, and applied to multiple KBP tasks. We demonstrated that our proposed framework is effective and suitable for collaborative development efforts, as well as useful in a teaching environment. Finally we present results that, while very modest, provide improvements an order of magnitude greater than our 2010 attempt~\cite{yu2010university}.


\bibliographystyle{acl}
\bibliography{kbp11}

\end{document}